%% file: main.tex
\documentclass[10pt,twocolumn,letterpaper]{article}

\usepackage{cvpr}              %

\usepackage{graphicx}
\usepackage{amsmath}
\usepackage{amssymb}
\usepackage{booktabs}

\usepackage{subcaption,siunitx,booktabs}
\usepackage{xcolor}
\usepackage{amsmath}
\usepackage{amssymb}
\usepackage{enumitem}

\usepackage{array}
\usepackage{caption}

\newcolumntype{L}[1]{>{\raggedright\let\newline\\\arraybackslash\hspace{0pt}}m{#1}}
\newcolumntype{C}[1]{>{\centering\let\newline\\\arraybackslash\hspace{0pt}}m{#1}}
\newcolumntype{R}[1]{>{\raggedleft\let\newline\\\arraybackslash\hspace{0pt}}m{#1}}
\usepackage{multirow}
\usepackage{graphicx}

\newcommand{\figref}[1]{Fig.~\ref{#1}}
\newcommand{\tabref}[1]{Tab.~\ref{#1}}

\newcommand{\equref}[1]{Equ.~(\ref{#1})}

\newcommand{\colwidthA}{1.0cm}
\newcommand{\colwidthB}{1.0cm}
\newcommand{\colwidthC}{1.0cm}

\usepackage[pagebackref,breaklinks,colorlinks]{hyperref}

\usepackage[capitalize]{cleveref}
\crefname{section}{Sec.}{Secs.}
\Crefname{section}{Section}{Sections}
\Crefname{table}{Table}{Tables}
\crefname{table}{Tab.}{Tabs.}

\def\cvprPaperID{7877} %
\def\confName{CVPR}
\def\confYear{2022}

\begin{document}

\title{
Locally Enhanced Self-Attention:\\
Combining Self-Attention and Convolution as Local and Context Terms}

\author{Chenglin Yang, Siyuan Qiao, Adam Kortylewski, Alan Yuille \\
Johns Hopkins University\\
{\tt\small \{chenglin.yang,siyuan.qiao,akortyl1,ayuille1\}@jhu.edu}
\and
}

\maketitle

\input{00_abstract}

\input{01_introduction}

\input{02_related_work}

\input{03_method}
\input{04_experiments}
\input{05_ablation_studies}

\input{06_conclusion}
\input{07_limitations}

{\small
\bibliographystyle{ieee_fullname}
\bibliography{egbib}
}

\end{document}

%% file: 00_abstract.tex
\begin{abstract}

Self-Attention has become prevalent in computer vision models. Inspired by fully connected Conditional Random Fields (CRFs), we decompose self-attention into local and context terms.
They correspond to the unary and binary terms in CRF and are implemented by attention mechanisms with projection matrices.
We observe that the unary terms only make small contributions to the outputs,
and meanwhile standard CNNs that rely solely on the unary terms achieve great performances on a variety of tasks.
Therefore, we propose Locally Enhanced Self-Attention (LESA), which enhances 
the unary term by incorporating it with convolutions, and utilizes a fusion module to dynamically couple the unary and binary operations.
In our experiments, we replace the self-attention modules with LESA. 
The results on ImageNet and COCO show the superiority of LESA over convolution and self-attention baselines for the tasks of image recognition, object detection, and instance segmentation. The code is made publicly available\footnote{\url{https://github.com/Chenglin-Yang/LESA}}.

\end{abstract}

%% file: 01_introduction.tex
\section{Introduction}

Self-Attention has made a great influence in the computer vision community recently.
It led to the emergence of fully attentional models~\cite{ramachandran2019stand,wang2020axial} and transformers~\cite{vaswani2017attention,dosovitskiy2020image,carion2020end}.
Importantly,
they show superior performances over traditional convolution neural networks
on a variety of tasks including classification, object detection, segmentation, and image completion~\cite{wang2020max,srinivas2021bottleneck,liu2021swin,wan2021high}. 
 
Despite its remarkable achievement, the understanding of self-attention remains limited. 
One of its advantages is 
overcoming the limitation of spatial distances on dependency modelling.
Originating from natural language processing, attention models the dependencies without regard to the distances among the words in the sequence, compared to LSTM~\cite{hochreiter1997long} and gated RNN~\cite{chung2014empirical}.
Being applied to vision models, attention aggregates the information globally among the pixels or patches~\cite{dosovitskiy2020image,wang2020axial}.
Similarly, compared to traditional convolutions, the features extracted by attention are no longer constrained by a local neighborhood.

\renewcommand{\colwidthA}{0.5cm}
\begin{table}[t]
 \centering
 \setlength{\tabcolsep}{0.0pt}
 \begin{tabular}{
        C{0.9cm}C{3cm}C{0.2cm}C{3cm}C{0.9cm}
 }
 {} & Image & {} & Convolution & {} \\[0.5em]
 {} & \includegraphics[width=3cm]{Figures/lesa_fig_1/img.jpg} & {} & \includegraphics[width=3cm]{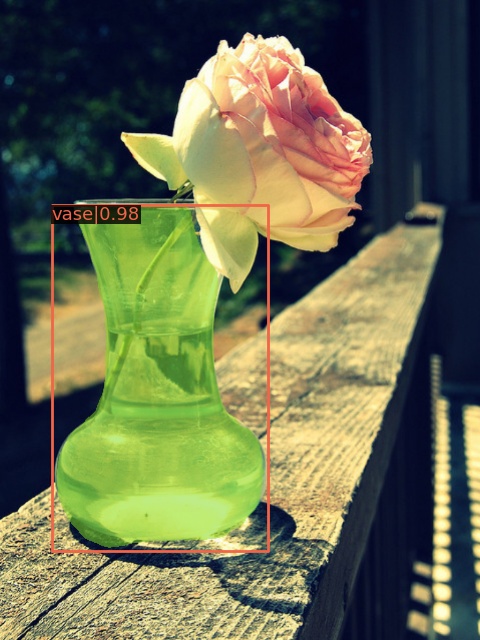} & {} \\[5.5em]
 {} & \includegraphics[width=3cm]{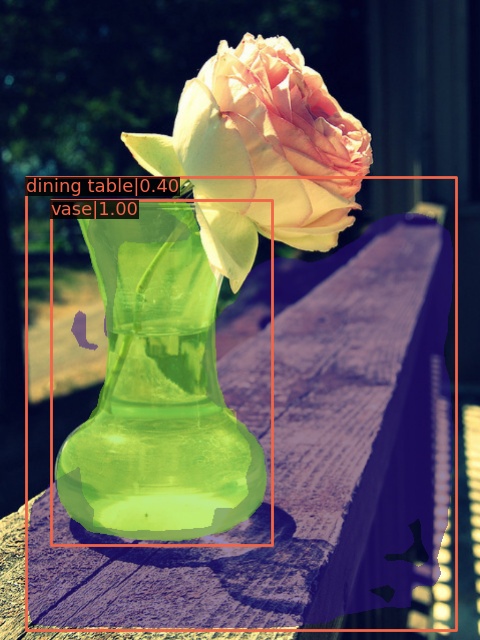} & {} & \includegraphics[width=3cm]{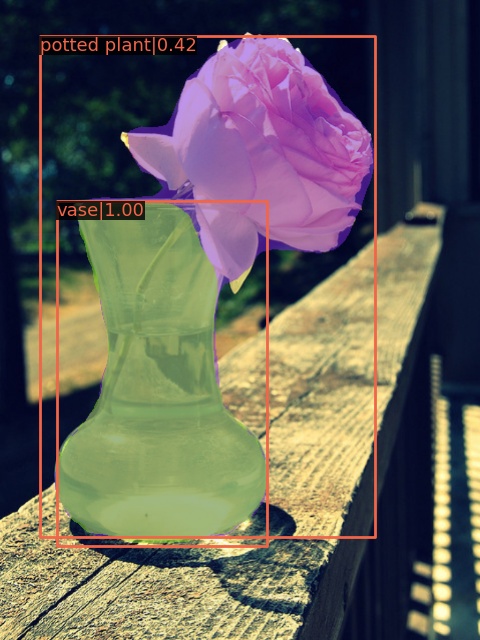} & {} \\[5.6em]
 {} & Self-Attention & {} & LESA & {} \\
 \end{tabular}
 \captionof{figure}{Effectiveness of Locally Enhanced Self-Attention (LESA) on COCO object detection and instance segmentation. All the outputs come from the Mask-RCNN models with WRN50 as the backbone. The training and testing processes are the same. The only difference is the spatial operations in the last two stages of WRN50. The comparisons are ensured fair. All the following visual examples are obtained by these models.}
 \label{fig: abs_lesa_figure}
\end{table}

\renewcommand{\colwidthA}{4.0cm}
\begin{table*}[!htp]
 \centering
 \small
 \setlength{\tabcolsep}{0.0pt}
 \begin{tabular}{C{\colwidthA}C{\colwidthA}C{\colwidthA}C{\colwidthA}}
    \multicolumn{4}{c}{\includegraphics[width=0.9\textwidth]{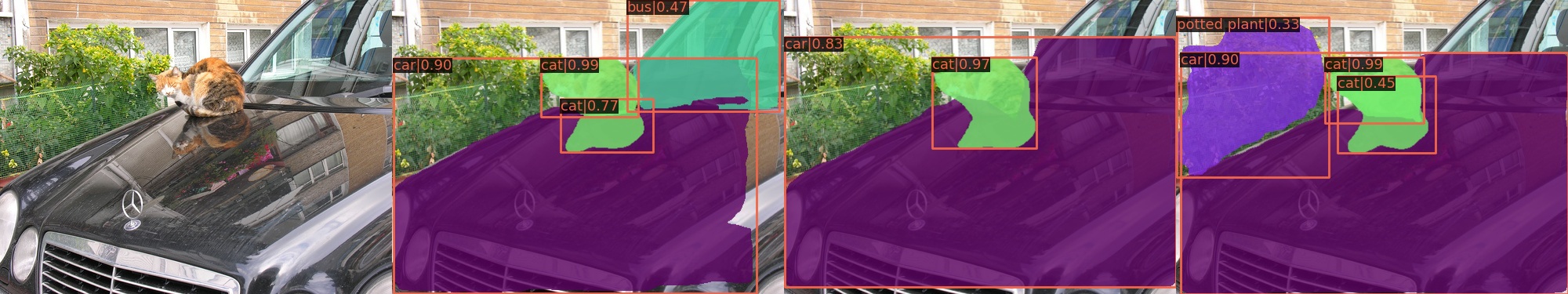}} \\
    \multicolumn{4}{c}{\includegraphics[width=0.9\textwidth]{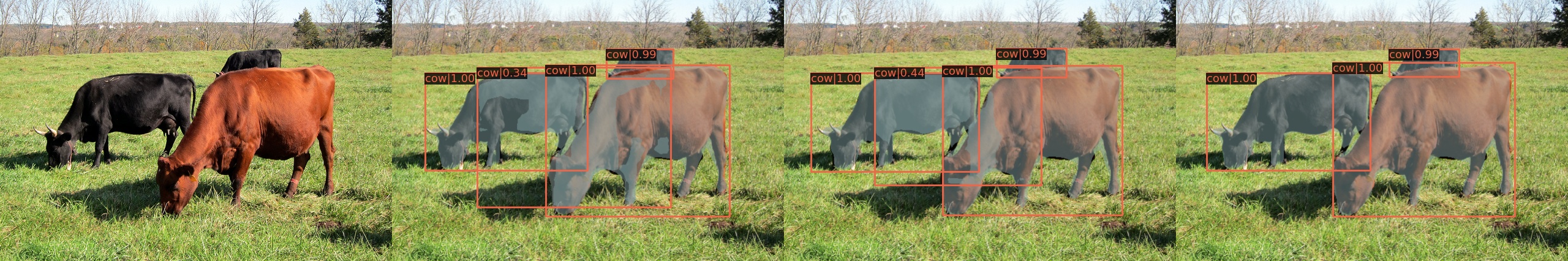}} \\
    Image & Convolution & Self-Attention & LESA \\
 \end{tabular}
 \captionof{figure}{Effectiveness of LESA on COCO object detection and instance segmentation. For detection (first row), the existence of local terms enables LESA more capable of detecting objects in detail and correctly. For segmentation (second row), the combination of local and context terms makes LESA generate masks with higher semantic consistency.}
 \label{fig: intro_lesa_figure}
\end{table*}

We argue that global aggregations in self-attention also bring problems because the aggregated features cannot clearly distinguish local and contextual cues.
We study this from the perspective of Conditional Random Fields (CRFs) and decompose it into local and context terms.
The unary (local) and binary (context) terms are based on the same building blocks of queries, keys, and values, and are calculated using the same projection matrices. 
We hypothesize that using the same building blocks for the local and context terms will cause problems,
which relates to the weaknesses of projections in self-attention pointed out by Dong \textit{et. al.}~\cite{dong2021attention}. 
They theoretically prove that the output of consecutive self-attention layers converges doubly exponentially to a rank-$1$ matrix and verify this degeneration in transformers empirically.
They also claim that skip connections can partially resolve the rank-collapse problem.
In our CRF analysis, the skip connections create the simplest local term which amounts to the identity mapping.
Skip connections alleviate the problem but we argue a local term with a stronger representation capacity needs to be designed.

\begin{figure*}[t]
    \centering
    \includegraphics[width=0.95\linewidth]{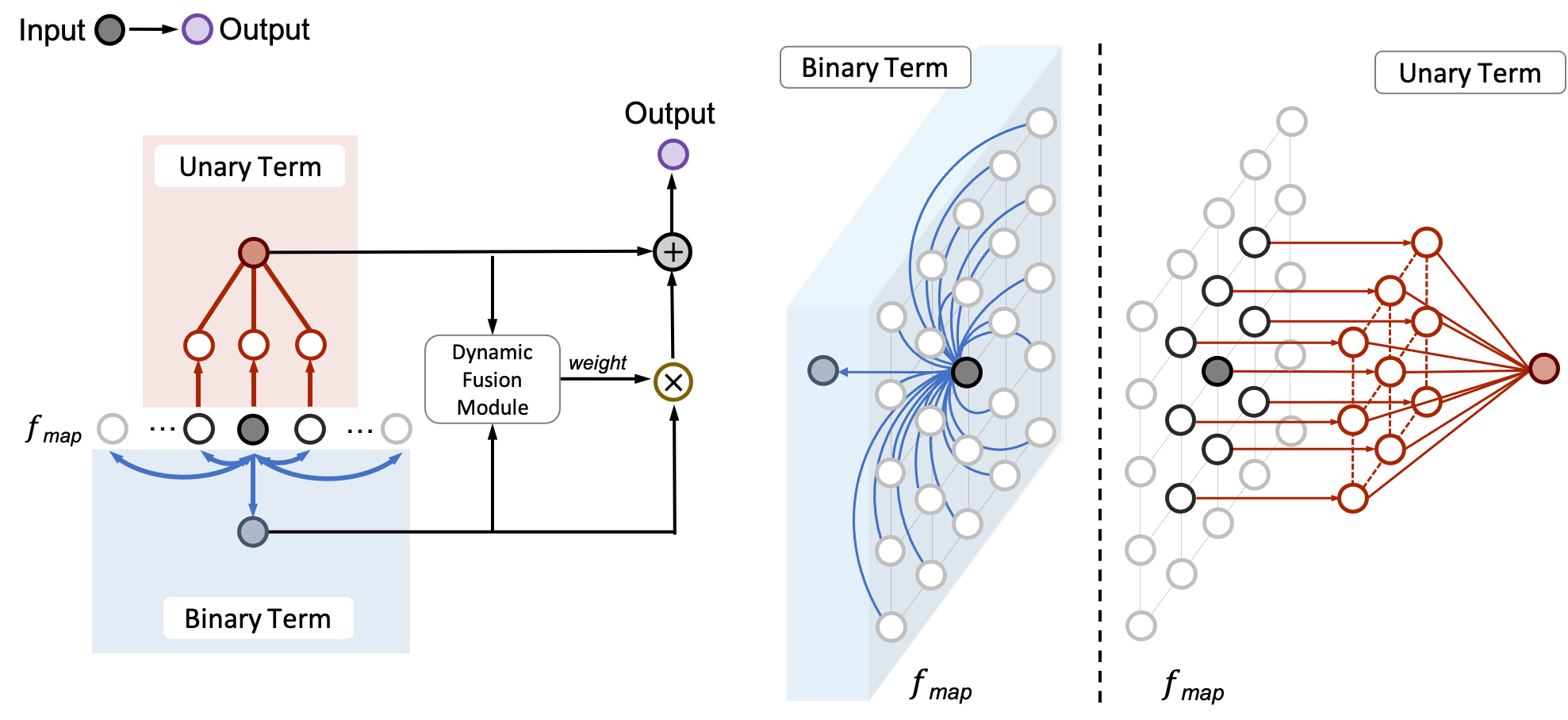}
    \caption{Visualizing Locally Enhanced Self-Attention (LESA) at one spatial location. The left part is the operation pipeline. $f_{map}$ is the feature map. The red and blue arrows represent the unary and binary operations. }
    \label{fig: LESA}
\end{figure*}

In this paper, we enhance the unary term by integrating it with convolutions and propose Locally Enhanced Self-Attention (LESA), which is visualized in \figref{fig: LESA}. To analyze self-attention from the perspective of CRFs,
let $x^{}$ be the input and $x^{+1}$ the output of one layer of self-attention.
Both of them are two-dimensional grid of nodes.
At spatial location $(i,j)$, the node $x^{+1}_{i,j}$ is connected to all the nodes $x^{}_{h,w}$ of the input. The binary term involves the computation on the edges $x_{h\neq i,w\neq j} \rightarrow x^{+1}_{i,j}$ while the unary term the computation on the edge $x_{h=i,w=j} \rightarrow x^{+1}_{i,j}$.
Intuitively, these two terms indicate the activation by looking at itself (local) and the others (context).
Through ablation study in \tabref{tab: res50_UB_contribution}, we find that the unary term is important for the performance but only contributing to the output less than $2\%$ computed by the softmax operation in attention.
Without the unary term, the feature extraction at $(i,j)$ entirely depends on interactions and losses the precise information of that pixel. 
The structure of the self-attention does not facilitate this unary operation. 
To address this issue, we enhance the unary term to $x_{(h,w) \in N(i,j)} \rightarrow x^{+1}_{i,j}$ where $N(i,j)$ indicate the pixels in neighborhood, and implement it as
a grouped convolution followed by a projection layer.

To couple the unary and binary terms, we propose a dynamic fusion mechanism. The simplest static ways would be to assign equal weights to them or by setting their weights to be hyper-parameters.
By contrast, we enable the model to allocate the weights on demand. Specifically, for each layer $l$ with the binary term, we multiply the binary term element-wisely by $\omega^{l}(x) \in \mathbb{R}^{H \times W \times C}$.
$\omega$ depends on the input $x$ and dynamically controls the weights of the binary terms to the unary terms for different layers $l$, spatial locations $H,W$, and feature channels $C$.

We study the performance of LESA for image classification, object detection, and instance segmentation. We replace the spatial convolutions with LESAs in the last two stages of ResNet~\cite{he2016deep} and its larger variant WRN~\cite{zagoruyko2016wide}.
Then, we use them equipped with FPN~\cite{lin2017feature} as the backbones in Mask-RCNN~\cite{he2017mask} to evaluate their performance for object detection and instance segmentation.
The challenging large-scale datasets ILSVRC2012~\cite{russakovsky2015imagenet} and COCO~\cite{lin2014microsoft} are used to train and evaluate the models.
The experiments demonstrate the superiority of LESA over the convolution and self-attention baselines.

To summarize, the main contributions of this work are:
\begin{itemize}[leftmargin=*]
    \item 
    Analyzing self-attention from the perspective of fully connected CRFs, we decompose it into a pair of local (unary) and context (binary) terms.
    We observe the unary terms make small contributions to the outputs.
    Inspired by the standard CNNs' focus on the local cues,
    we propose to enhance the unary term by incorporating it with convolutions.
    
    \item We propose a dynamic fusion module to couple the unary and binary terms adaptively.
    Their relative weights are adjusted as needed, depending on specific inputs, spatial locations, and feature channels.
    \item We implement Locally Enhanced Self-Attention (LESA) for vision tasks. 
    Our experiments on the challenging datasets, ImageNet and COCO, demonstrate that the LESA is superior to the convolution and self-attention baselines.
    Especially for object detection and instance segmentation where local features are particularly important, LESA achieves significant improvements.
\end{itemize}

%% file: 02_related_work.tex
\section{Related Work}

\subsection{Convolution}

Convolutional Neural Networks (CNNs) have become the dominant models in computer vision in the last decade. AlexNet~\cite{krizhevsky2012imagenet} shows considerable improvement over the models based on hand-crafted features~\cite{perronnin2010improving,sanchez2011high}, and opens the door to the age of deep neural networks. Lots of efforts have been made to increase the width and depth and to improve the architecture and efficiency of CNNs in the pursuit of performance. They include the designs of VGG~\cite{simonyan2014very}, GoogleNet~\cite{szegedy2015going}, ResNet~\cite{he2016deep}, WRN~\cite{zagoruyko2016wide}, ResNeXt~\cite{xie2017aggregated}, DenseNet~\cite{huang2017densely}, SENet~\cite{hu2018squeeze}, MobileNet~\cite{howard2017mobilenets}, EfficientNet~\cite{tan2019efficientnet}, etc. Through this process, the convolution layers are also being developed, leading to the grouped convolutions~\cite{xie2017aggregated}, depth-wise separable convolutions~\cite{chollet2017xception}, deformable convolutions~\cite{dai2017deformable,zhu2019deformable}, atrous convolutions~\cite{chen2014semantic,papandreou2015modeling} and switchable atrous convolutions~\cite{qiao2020detectors,chen2020scaling}. 

\subsection{Self-Attention}

The impact of self-attention on vision community is becoming greater.
Self-attention is originally proposed in approaches of natural machine translation~\cite{bahdanau2014neural}. It enables the encoder-decoder model to adaptively find the useful information according to contents from a variable length sentence. In computer vision, non-local neural networks~\cite{wang2018non} show that self-attention is an instantiation of non-local means~\cite{buades2005non}, and use it to capture long-range dependencies to augment CNNs for tasks including video classification and object detection. 
$A^{2}$-Net~\cite{chen20182} employs a variant of non-local means and shows performance improvement on image classification.
Recently, fully attentional methods~\cite{ramachandran2019stand,hu2019local,zhao2020exploring} which replaces all the spatial convolutions with self-attention in the deep networks are proposed with stronger performances than CNNs. Axial attention~\cite{wang2020axial} factorizes the 2D self-attention into two 1D consecutive self-attentions which reduces the computation complexity and enables the self-attention layer to have a global kernel. Self-attention also promotes the generation of transformers~\cite{vaswani2017attention,dosovitskiy2020image,touvron2020training,carion2020end,wu2020lite,liu2021swin}. 
BotNet~\cite{srinivas2021bottleneck} relates the transformer block with the fully attentional version of bottleneck block in ResNet~\cite{he2016deep}. 

\subsection{Combining Self-Attention and Convolution}

There are four categories of methods to combine the self-attention and convolution. \textit{Approaches in different categories can be used together.} 

The first one is using depth-wise convolution to replace the position embedding layer in self-attention, including CVPT~\cite{chu2021conditional} and CoaT~\cite{xu2021co}. 
The second one is the serial connection with convolution before attention. CvT~\cite{wu2021cvt} applies convolutions before calculating the query, key and value in self-attention.
The third one is the serial connection with attention before convolution. Convolutions are applied after the self-attention layer, including LocalViT~\cite{li2021localvit} and PVTv2~\cite{wang2021pvtv2}.
The final one is parallel connection. Attention Augmentation~\cite{bello2019attention} augments the convolution features with attention features. It is intended to incorporate the long-range connections into the convolutions. Our approach, LESA falls into this category. Different from Attention Augmentation, it enhances the local term in self-attention.

%% file: 03_method.tex
\section{LESA: Locally Enhanced Self Attention}

\subsection{Decomposition of Self-Attention}
\label{sec: dompose self-attention}

We decompose the self-attention into local and context terms.
Let $x \in \mathbb{R}^{d_{in}\times H \times W}$ be the input, where $d_{in}$ is the feature channels and $H, W$ are the height and width in spatial dimensions.
In this case, each pixel is connected with all the other pixels in the computation.
We consider the all-to-all self-attention since it has been adopted as a building layer and shows superior performance~\cite{wang2020axial, srinivas2021bottleneck}.
Specifically, we can write the formula of self-attention as:
\begin{align}
\begin{split}
        x_{i,j}^{l+1} = \sum_{(h, w) = (1, 1)}^{(H, W)} \text{softmax} (q_{i,j}^\top k_{h,w} + q_{i,j}^{\top} r_{(i,j) \rightarrow (h,w)} ) v_{h,w}
\end{split}
\end{align}
where $i, j$ and $h, w$ represent the spatial locations of the pixel and $l$ specifies the layer index. $q, k, v = xW_{q}, xW_{k}, xW_{v}$ are the query, key and value which are obtained by applying three different $1 \times 1$ convolutions on $x^{l}$. $W_{q}, W_{k} \in \mathbb{R}^{d_{in} \times d_{qk}}$ and $W_{v} \in \mathbb{R}^{d_{in} \times d_{out}}$ are learnable parameters, where $d_{qk}, d_{out}$ are intermediate and output channels.
$r_{(i,j)\rightarrow (h,w)}$ is the relative position embedding, and for simplicity we will use the notation $r_{h,w}$.
This formula shows the activation $x_{i,j}^{l+1}$ integrates the information conveyed by all the pixels $x_{i, j}^{l}$. To comprehend this operation, we decompose the information flow and reformulate the equation as the combination of local term and context term:
\begin{equation}
    \begin{split}
    x_{i,j}^{l+1} = \quad &\text{S}_{H,W}^{i,j} (q_{i,j}^\top k_{i,j} + q_{i,j}^\top r_{i,j} ) v_{i,j} + \\
    & \sum_{(h, w) \neq (i, j)}^{(H, W)} \text{S}_{H,W}^{h,w} (q_{i,j}^\top k_{h,w} + q_{i,j}^\top r_{h,w} ) v_{h,w} \label{eq: decompostion} \\
    \end{split}
\end{equation}
\begin{equation}
    \text{S}_{H,W}^{h,w}(z) = \frac{e^{z_{h,w}}}{\sum_{(m,n)}^{(H,W)}e^{z_{m,n}}}
\end{equation}
For the spatial location $(i, j)$, the first local term computes activation by looking at itself, while the second context term computes activation by looking at others. The softmax generates the weights of contribution. Through this decomposition, we can interpret self-attention as a double-source feature extractor, which consists of a pair of unary and binary terms.
Unary and binary terms are computed by the queries, keys, and values $q, k, v$ at different spatial locations with \textit{shared} projection matrices $W_{q}, W_{k}, W_{v}$.
Consequently, the outputs \textit{entangle} the local and context features.

We perform an ablation study to investigate the contribution of these two terms. Specifically, we take a ResNet50~\cite{he2016deep} and replace the convolution layers of its last two stages with self-attention. The model is trained from scratch on ImageNet~\cite{russakovsky2015imagenet}. 
During inference, we track the softmax operations of all self-attention layers and obtain the weights for the unary and binary terms,  $\text{S}_{H,W}^{h,w}$ and $\sum_{(h, w) \neq (i, j)}^{(H, W)} \text{S}_{H,W}^{h,w}$ whose summation equals to $1$. By averaging them across all layers, we obtain the weighted contributions of these two terms. Then, we ablate the unary in the evaluation phase.
The results are shown in \tabref{tab: res50_UB_contribution}. We can observe that self-attention is mainly contributed by the binary operations, but the unary term is also important.
Although the weights of unary terms only take less than $2\%$, the removal of which causes $7.56\%$ drop of accuracy or $35\%$ relative increase on the error rate.
When analyzing the self-attention by this decomposition, unary term plays a significant role, but most of the computations and focuses are given to binary operations.
\begin{table}[t]
  \centering
  \begin{tabular}{ccc}
    \toprule
    Methods     & Top-1 Err. (\%)    & Weight Pct. (\%)\\
    \midrule
    SA    & $21.31$   & $100.00$  \\
    SA - unary term            & $28.87$   & $98.03$   \\
    \bottomrule
  \end{tabular}
  \caption{Contributions of the unary term in Self-Attention (SA). We replace the spatial convolutions in the $3$rd and $4$th stages of ResNet50~\cite{he2016deep} with self-attention. By tracking the softmax operations, we record the weights of the unary and binary terms $\text{S}_{H,W}^{h,w}$ and $\sum_{(h, w) \neq (i, j)}^{(H, W)} \text{S}_{H,W}^{h,w}$ in \equref{eq: decompostion}. They add up to $1$ at each layer. The weight percentage is the average across all the layers. We observe that
  the unary term is important. 
  The removal of unary terms whose weight percentage is less than $2\%$ increases the error rate by $7.56\%$ (or $>35\%$ relative increase).
  }
  \label{tab: res50_UB_contribution}
\end{table}

\subsection{Locally Enhanced Self-Attention
}

Local and context terms have been long used in formulating the graphical models for vision tasks, 
such as image denoising, segmentation, and surface reconstruction~\cite{prince2012computer}.
The fully connected Conditional Random Fields (CRFs) 
have been introduced on top of the deep networks for semantic segmentation~\cite{chen2017deeplab}. It aims at coupling the recognition capacity and localization accuracy, and achieves excellent performance. 
For a grid of pixels in the form of a graph $G = (V, E)$, the energy to be minimized for the CRF is defined by:
\begin{align}
    \phi_c &= \sum_{x \in V} \Phi_U(x) + \sum_{(x_u, x_v) \in E} \Phi_B(x_u, x_v) 
\end{align}
where $u, v$ indicate the different vertices in $V$. The unary term is $\Phi_U (x) = -\log (P(x))$ where $P(x)$ is the probability of assigning $x$ the ground truth label by the model. The binary term is $\Phi_B (x) = \pi((x_u, p_u), (x_v, p_v)) + \pi(p_u, p_v) $ where $x$ and $p$ are the contents and spatial positions.
$\pi$ is the probability density function to measure the similarity of two values, which can be chosen as Gaussian. 

The unary term is utilized for recognition while the binary term for spatial and content interactions.
Inspired by these and our decomposition analysis, we propose Locally Enhanced Self-Attention (LESA).
It contains a unary term incorporated with convolutions, and a binary term for feature interactions.
Locally Enhanced Self-Attention is defined by
\begin{gather}
\label{eq: dual-view descriptor}
     x_{i,j}^{l+1} = m_{i, j} + \quad \omega_{i,j} \sum_{(h, w) = (1, 1) }^{(H, W)}  \text{S}_{H,W}^{h,w} q_{i,j}^\top k_{h,w} v_{h,w} \\
     \quad m = xW_{g}^{k}W_{1}^{1}
     \label{eq: unary term}
\end{gather}
where $\omega_{i,j}$ is the weight that is shown in~\equref{eq: omega}.
$m$
is the local term obtained by applying two consecutive convolutions.
$W_{g}^{k} \in \mathbb{R}^{d_{in} \times d_{out}}$ is a learnable matrix where $k$ and $g$ represent the spatial extent and group number of the convolution. $W_{1}^{1} \in \mathbb{R}^{d_{out} \times d_{out}}$ is a learnable projection matrix representing $1 \times 1$ convolution. By this design, the multi-head mechanism is integrated. 
$m_{i,j}$ is the unary activation at spatial location $(i,j)$.
This formulation of LESA also enables us to change $W_{g}^{k}$ to deformable convolution~\cite{dai2017deformable} for the tasks of object detection and instance segmentation.
Self-attention focuses on the binary operations.
We use it as the context term to model the feature interactions with relative spatial relationships among all possible pairs of pixels. 

\begin{table}[!btp]
  \centering
  \small
  \begin{tabular}{c|c|c|cc|cc}
    \toprule
    \multirow{2}{*}{Models} & \multirow{2}{*}{Ops.} & \multirow{1}{*}{Pms} & \multicolumn{2}{c|}{Acc.(\%)} & \multicolumn{2}{c}{Weights (\%)}  \\
    {} & {} & {(M)} & T-1 & T-5 & Unary & Binary \\
    \midrule
    \midrule
    R50 & Conv.     & $25.6$ & $76.1$ & $92.9$  & $100$ & $0.0$  \\
    R50 & SA  & $18.1$ & $78.7$ & $94.2$  & $2.0$ & $98.0$  \\
    R50 & LESA  & $23.8$ & $79.6$ & $94.8$  & $67.0$ & $33.0$  \\
    \midrule
    WR50 & Conv.             & $68.9$ & $78.5$ & $94.1$  & $100$ & $0.0$  \\
    WR50 & SA          & $37.7$ & $79.7$ & $94.7$  & $2.5$ & $97.5$  \\
    WR50 & LESA          & $60.6$ & $80.2$ & $95.1$  & $66.9$ & $33.2$  \\
    \bottomrule
  \end{tabular}
  \caption{
  ImageNet classification results. We replace the spatial convolutions in the $3$rd and $4$th stages of ResNet~\cite{he2016deep} and WRN~\cite{zagoruyko2016wide} with self-attention and LESA. We employ the position embedding method used in~\cite{wang2020axial}. For self-attention, the weight percentages are obtained by tracking softmax operations which is the same as \tabref{tab: res50_UB_contribution}. For LESA, the $\omega$ in \equref{eq: omega}, the output of the dynamic fusion module is tracked. $\frac{1}{1+\omega}$ and $\frac{\omega}{1+\omega}$ are the weights for the unary and binary terms in LESA, which are then averaged across the spatial locations, feature channels, and layers as the final weight percentage. 
  We can observe that compared with self-attention, LESA achieves a more balanced utilization of both the unary and binary terms.
  In both the top-1 and top-5 accuracy, LESA outperforms other baselines. 
  \textit{LESA has a more significant improvement in object detection where the local cues are particularly important.
  }
  }
  \label{tab: ImageNet}
\end{table}

\subsection{Dynamic Fusion of the Unary and Binary Terms}
\label{sec: omega}

Adding the unary and binary terms is a static way of merging the two terms with equal weights.
A more flexible strategy is to allocate the weights on demand under different circumstances. For example, in object detection, the locality of pixel dependencies is more important than the context when detecting multiple small objects in an image.
We achieve a dynamic control by multiplying the binary term by $\omega$ and adaptively adjusting the relative weights of the two terms, which is shown in \equref{eq: dual-view descriptor}.
Specifically, we can write the formula of $\omega_{i,j}$ as:
\begin{align}
\label{eq: omega}
    \omega_{i,j} = \text{sigmoid} \Big( F \big(  m_{i, j} \mathbin\Vert \sum_{(h, w) = (1, 1) }^{(H, W)} \text{S}_{H,W}^{h,w} q_{i,j}^\top k_{h,w}v_{h,w} \big) \Big)
\end{align}
where $\omega \in \mathbb{R}^{d_{out} \times H \times W}$ and $\omega_{i,j}$ corresponds to one spatial location.
$F: \mathbb{R}^{2d_{out}\times H \times W} \mapsto \mathbb{R}^{d_{out}\times H \times W}$ is a function. Sigmoid operation is performed element-wisely on the logits given by $F$, making $\omega$ range from $0$ to $1$. Regarding $F$, we design it as a three-layer perceptron and adopt the pre-activation design~\cite{he2016identity}. Concretely, together with sigmoid we can represent the pipeline as
$F + \text{sigmoid} : \text{BN} \rightarrow \text{relu} \rightarrow fc_{1} \rightarrow \text{BN} \rightarrow \text{relu} \rightarrow fc_{2} \rightarrow \text{BN} \rightarrow \text{sigmoid} $
where BN is batch normalization layer~\cite{ioffe2015batch}, and $fc_{1}: \mathbb{R}^{2d_{out}} \mapsto \mathbb{R}^{d_{out}}$, $fc_{2}: \mathbb{R}^{d_{out}} \mapsto \mathbb{R}^{d_{out}}$ are two fully connected layers. The position embedding is omitted in Equ.~\ref{eq: dual-view descriptor} and \ref{eq: omega} for simplicity. 
In our design, 
$\omega$ depends on the contents of the unary and binary terms and controls their relative weights at different spatial locations and in different feature channels. 

%% file: 04_experiments.tex
\section{Experiments}

\subsection{ImageNet Classification}
\label{sec: imagenet}

\renewcommand{\colwidthB}{1.00cm}
\renewcommand{\colwidthC}{0.70cm}
\begin{table*}[t]
  \centering
  \small
  \begin{tabular}{c|c|c|ccc|ccc|ccc|ccc}
    \toprule
    \multirow{2}{*}{BB.} & \multirow{2}{*}{Ops.} & \multirow{2}{*}{Epo.} & \multicolumn{6}{c|}{Object Detection} & \multicolumn{6}{c}{Intancecs Segmentation} \\
    \cline{4-15}
    {} & {} & {}  & $\text{AP}^{\text{b}}_{\text{}}$    & $\text{AP}^{\text{}}_{\text{50}}$  & $\text{AP}^{\text{}}_{\text{75}}$ &  $\text{AP}^{\text{}}_{\text{S}}$    & $\text{AP}^{\text{}}_{\text{M}}$  & $\text{AP}^{\text{}}_{\text{L}}$ & $\text{AP}^{\text{m}}_{\text{}}$    & $\text{AP}^{\text{}}_{\text{50}}$  & $\text{AP}^{\text{}}_{\text{75}}$ &  $\text{AP}^{\text{}}_{\text{S}}$    & $\text{AP}^{\text{}}_{\text{M}}$  & $\text{AP}^{\text{}}_{\text{L}}$  \\
    \midrule
    R50  &   Conv.     & $12$ & $37.8$ & $58.6$ & $41.0$ & $22.1$ & $41.1$ & $48.3$ & $34.5$ & $55.5$ & $36.7$ & $18.6$ & $37.5$ & $45.8$    \\
    R50  &   SA  & $12$ & $38.5$ & $60.0$ & $41.6$ & $23.7$ & $41.4$ & $49.5$ & $34.7$ & $56.7$ & $36.6$ & $19.7$ & $37.5$ & $46.4$ \\
    R50  &   LESA  & $12$ & $40.3$ & $61.2$ & $43.6$ & $24.7$ & $43.3$ & $52.0$ & $36.5$ & $58.3$ & $38.9$ & $21.0$ & $39.5$ & $48.9$    \\
    \midrule
    WR50  &   Conv.     & $12$ & $40.9$ & $61.8$ & $44.8$ & $24.2$ & $44.7$ & $52.5$ & $36.9$ & $58.6$ & $39.6$ & $20.3$ & $40.5$ & $49.0$    \\
    WR50  &   SA  & $12$ & $42.0$ & $63.2$ & $45.6$ & $25.4$ & $45.0$ & $54.6$ & $37.6$ & $60.1$ & $40.1$ & $21.1$ & $40.5$ & $50.4$    \\
    WR50  &   LESA  & $12$ & $44.8$ & $66.1$ & $49.1$ & $27.8$ & $48.3$ & $58.4$  & $39.9$ & $62.9$ & $42.8$ & $22.7$ & $43.1$ & $54.0$    \\
    \midrule
    R50  &   Dconv.           & $20$ & $43.0$ & $64.0$ & $47.3$ & $25.8$ & $46.4$ & $56.4$ & $38.6$ & $61.0$ & $41.2$ & $21.4$ & $41.8$ & $52.4$    \\
    R50  &   LESA  & $20$ & $44.0$ & $64.8$ & $48.0$ & $26.3$ & $47.9$ & $57.3$  & $39.2$ & $61.8$ & $41.9$ & $21.9$ & $42.8$ & $52.5$    \\
    \midrule
    WR50  &   Dconv.           & $20$ & $44.7$ & $65.5$ & $49.1$ & $27.0$ & $48.2$ & $59.3$  & $39.8$ & $62.5$ & $42.7$ & $22.6$ & $43.1$ & $54.7$    \\
    WR50  &   LESA  & $20$ & $46.3$ & $67.5$ & $50.7$ & $28.8$ & $49.9$ & $60.2$ & $40.9$ & $64.3$ & $43.9$ & $23.5$ & $44.5$ & $54.9$    \\
    \midrule
    WR50  &   HTC Dconv.           & $20$ & $47.4$ & $66.7$ & $51.7$ & $30.0$ & $50.9$ & $61.8$ & $41.9$ & $64.0$ & $45.4$ & $24.2$ & $45.3$ & $57.1$    \\
    WR50  &   LESA  & $20$ & $48.8$ & $68.4$ & $53.3$ & $31.7$ & $52.3$ & $63.1$ & $43.0$ & $65.7$ & $46.4$ & $25.8$ & $46.4$ & $57.8$    \\
    \midrule
    WR50  &   HTC Dconv$^{\text{h}}$  & $20$ & $48.5$ & $67.8$ & $52.9$ & $31.3$ & $52.0$ & $62.3$ & $42.8$ & $65.6$ & $46.5$ & $25.6$ & $46.1$ & $57.1$    \\
    WR50  &   LESA\_H  & $20$ & $50.1$ & $69.7$ & $54.4$ & $32.3$ & $53.6$ & $65.0$ & $43.9$ & $67.1$ & $47.5$ & $26.6$ & $47.1$ & $59.1$    \\
    \bottomrule
  \end{tabular}
  \caption{
  COCO object detection and instance segmentation on \texttt{val2017}. We use Mask-RCNN~\cite{he2017mask} and HTC~\cite{chen2019hybrid} frameworks and employ FPN~\cite{lin2017feature} on the backbones ResNet~\cite{he2016deep} and WRN~\cite{zagoruyko2016wide}. Donv. stands for deformable convolutions.
  We adopt two standard training schedules that have $12$ and $20$ epochs.
  The learning rate is adjusted $10$ times smaller after $8, 11$ epochs and $16, 19$ epochs, respectively. The images are resized to $1024 \times 1024$ by default, and the superscript $^{\text{h}}$ (higher resolution) indicates that they are resized to $1280 \times 1280$.
  We can observe that LESA outperforms the convolution, self-attention, and deformable convolution baselines in all the experiments.
  }
  \label{tab: COCO_detection_val}
\end{table*}

\renewcommand{\colwidthB}{1.00cm}
\renewcommand{\colwidthC}{0.80cm}
\begin{table*}[h]
  \centering
  \small
  \begin{tabular}{c|c|c|ccc|ccc|ccc|ccc}
    \toprule
    \multirow{2}{*}{BB.} & \multirow{2}{*}{Ops.} & \multirow{2}{*}{Epo.} & \multicolumn{6}{c|}{Object Detection} & \multicolumn{6}{c}{Intancecs Segmentation} \\
    \cline{4-15}
    {} & {} &  {}  & $\text{AP}^{\text{b}}_{\text{}}$    & $\text{AP}^{\text{}}_{\text{50}}$  & $\text{AP}^{\text{}}_{\text{75}}$ &  $\text{AP}^{\text{}}_{\text{S}}$  & $\text{AP}^{\text{}}_{\text{M}}$  & $\text{AP}^{\text{}}_{\text{L}}$  & $\text{AP}^{\text{m}}_{\text{}}$    & $\text{AP}^{\text{}}_{\text{50}}$  & $\text{AP}^{\text{}}_{\text{75}}$ &  $\text{AP}^{\text{}}_{\text{S}}$  & $\text{AP}^{\text{}}_{\text{M}}$  & $\text{AP}^{\text{}}_{\text{L}}$ \\
    \midrule
    R50  &   LESA  & $12$ & $40.5$ & $61.7$ & $44.3$ & $24.2$ & $42.4$ & $50.1$  & $36.6$ & $58.7$ & $39.0$ & $20.6$ & $38.3$ & $46.9$    \\
    \midrule
    WR50  &   LESA  & $12$ & $44.8$ & $66.4$ & $49.2$ & $26.9$ & $47.8$ & $56.0$ & $40.0$ & $63.1$ & $43.1$ & $22.7$ & $42.6$ & $51.8$    \\
    \midrule
    R50  &   LESA  & $20$ & $44.2$ & $65.5$ & $48.3$ & $26.4$ & $46.9$ & $55.5$  & $39.6$ & $62.4$ & $42.6$ & $22.5$ & $42.0$ & $51.1$    \\
    \midrule
    WR50  &   LESA  & $20$ & $46.5$ & $67.7$ & $51.0$ & $28.4$ & $49.4$ & $58.3$ & $41.2$ & $64.6$ & $44.5$ & $23.9$ & $43.7$ & $53.4$    \\
    \midrule
    WR50  &   HTC LESA  & $20$ & $49.0$ & $68.8$ & $53.4$ & $30.3$ & $51.8$ & $61.7$  & $43.4$ & $66.4$ & $47.1$ & $25.2$ & $45.9$ & $56.6$    \\
    \midrule
    WR50  &   HTC LESA$^{\text{h}}$  & $20$ & $50.5$ & $70.2$ & $54.9$ & $31.8$ & $53.2$ & $63.3$ & $44.4$ & $67.7$ & $48.2$ & $26.6$ & $46.7$ & $57.5$    \\
    \bottomrule
  \end{tabular}
  \caption{
  COCO object detection and instance segmentation on \texttt{test-dev2017} for the models in~\tabref{tab: COCO_detection_val}.
  }
  \label{tab: COCO_detection_test}
\end{table*}

\noindent$\bullet$\quad{\bf Settings}
We perform image classification experiments on ILSVRC2012~\cite{russakovsky2015imagenet} which is a popular subset of the ImageNet database~\cite{deng2009imagenet}. There are $1.3\mathrm{M}$ images in the training set and $50\mathrm{K}$ images in the validation set. In total, it includes $1\rm{,}000$ object classes.

ResNet~\cite{he2016deep}, a family of canonical models or backbones for vision tasks, and its larger variant WRN~\cite{zagoruyko2016wide} are used to study LESA. There are $4$ stages in ResNet and each one is formed by a series of bottleneck blocks. ResNet50 can be represented by the bottleneck numbers $\{3,4,6,3\}$. 
We replace the conv$3\times 3$ in the bottleneck with the self-attention and LESA. 
The kernel channels of these conv$3 \times 3$ in WRN are twice as large as those in ResNet. 

We perform the replacement in the last two stages, which is enough to show the advantages of LESA.
For convolution baselines, we use Torchvision official models~\cite{paszke2019pytorch}. For self-attention baselines and LESA, we set head number $8$ for both of them and train the models from scratch. We set the stride of the last stage to be $1$ following~\cite{srinivas2021bottleneck}. We keep the first bottleneck in stage $3$ unchanged which has the stride $2$ convolution.
We employ a canonical training scheme with $5$ linear warm-up and $90$ training epochs with a batch size $128$. Following ~\cite{ramachandran2019stand,wang2020axial}, we employ SGD with Nesterov momentum~\cite{nesterov1983method,sutskever2013importance} and cosine annealing learning rate initialized as $0.05$. 
The experiments are performed on $4$ NVIDIA TITAN XP graphics cards.

\noindent$\bullet$\quad{\bf Results} 
The results are summarized in \tabref{tab: ImageNet}. For both the top-$1$ and top-$5$ accuracy, LESA surpasses the convolution and self-attention baselines. Our dynamic fusion module controls the binary term using $\omega$ in \equref{eq: omega}, and thus the weights for the unary and binary terms are $\frac{1}{1+\omega}$ and $\frac{\omega}{1+\omega}$, respectively. As $w$ is dependent on the inputs, spatial locations, and feature channels, we average the weights across them in our records. 
In self-attention, the weights are calculated by softmax operations as used in \tabref{tab: res50_UB_contribution}.
It is observed that the weight distribution in self-attention 
are imbalanced. The unary term only has a weight percentage less than $3\%$, more than $32$ times smaller than the binary term's. While for LESA, 
their weight percentages are about $67\%$ and $33\%$, respectively. In the tasks of object detection where local cues are particularly important, LESA shows better improvement, which is shown in Tab. \ref{tab: COCO_detection_val} and \ref{tab: COCO_detection_test}.

\subsection{COCO Object Detection and Instance Segmentation.}
\label{sec: coco}

\noindent$\bullet$\quad{\bf Settings}
We perform object detection experiments on COCO dataset~\cite{lin2014microsoft} and use the 2017 dataset splits. There are $118\mathrm{K}$ images in the training set and $5\mathrm{K}$ images in the validation set. In total there are $80$ object categories and on average each image contains $3.5$ categories and $7.7$ instances.

The widely used Mask-RCNN~\cite{he2017mask} and HTC~\cite{chen2019hybrid} with the backbones equipped with FPN~\cite{lin2017feature} are used to study LESA for object detection and instance segmentation. We use mmdetection~\cite{chen2019mmdetection} as the codebase. The ImageNet pre-trained checkpoints are utilized to initialize the backbones. 
There are $5$ stages in the ResNet-FPN and the output strides are $\{4, 8, 16, 32, 64\}$. 
We replace the spatial convolutions in the $3$rd and $4$th stages.
The images are resized to $1024 \times 1024$ and $1280 \times 1280$ in the experiments.
Since the image size in classification is $224 \times 224$, we initialize new position embedding layers used in~\cite{shaw2018self,ramachandran2019stand}. 
For training, we employ the $\boldsymbol{1 \times}$ and $\boldsymbol{20e}$ schedules. The total epochs and epochs after which the learning rate is multiplied by $0.1$ are $\{12, 8, 11\}$ and $\{24, 16, 19\}$, respectively.
For the HTC framework, we employ multi-scale training for both the baseline and our method: with $0.5$ probability that both sides of the image are resized to a scale uniformly chosen from $512$ to $1024$, and with another $0.5$ probability to a scale that is uniformly sampled from $1024$ to $2048$.
Mask-RCNN does not use multi-scale training.

We also study adopting the deformable unary terms in LESA. Specifically, we replace $W_{g}^{k}$ in \equref{eq: unary term} to deformable convolutions~\cite{dai2017deformable}. We set the group number of offsets as $1$.
Following the standard setting~\cite{qiao2020detectors}, the convolutions in the $2$nd stage in both the baselines and our models are also replaced with deformable convolutions.
Our experiments with Mask-RCNN framework are performed on $4$ NVIDIA TITAN XP graphics cards and those with HTC framework on $4$ TITAN RTX graphics cards.

\noindent$\bullet$\quad{\bf Results}
The results are summarized in Tab.~\ref{tab: COCO_detection_val} and~\ref{tab: COCO_detection_test}. 
We use the same testing pipeline for \texttt{val2017} and \texttt{test-dev2017}. 
LESA provides the best bounding box mAP and mask mAP for all the small, medium, and large objects compared with convolution, self-attention, and deformable convolution baselines in all scenarios.

\label{sec: exp}

\renewcommand{\colwidthA}{2.6cm}
\begin{table}[!htp]
 \centering
 \small
 \setlength{\tabcolsep}{0.0pt}
 \begin{tabular}{C{\colwidthA}C{\colwidthA}C{\colwidthA}C{\colwidthA}}
        \multicolumn{3}{c}{\includegraphics[width=0.45\textwidth]{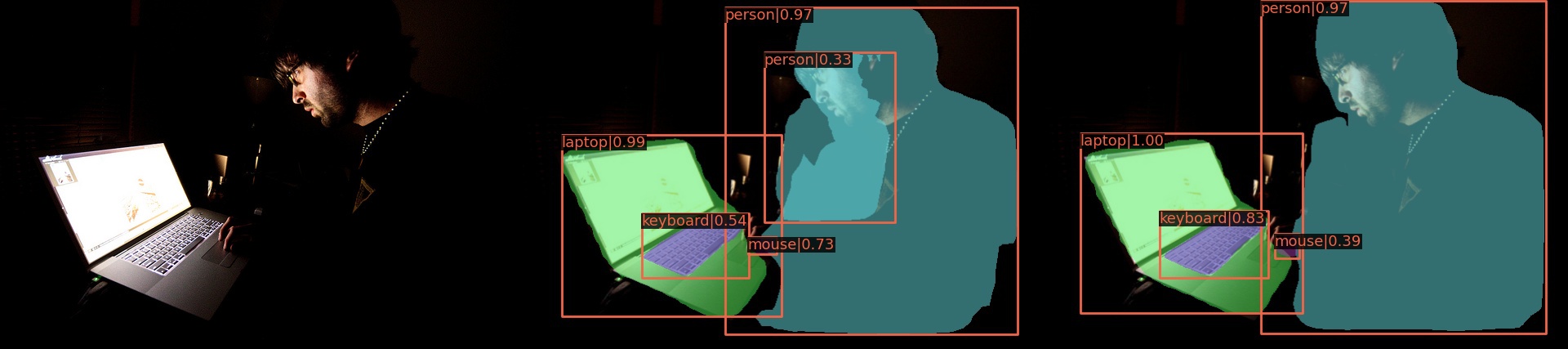}} \\
    \multicolumn{3}{c}{\includegraphics[width=0.45\textwidth]{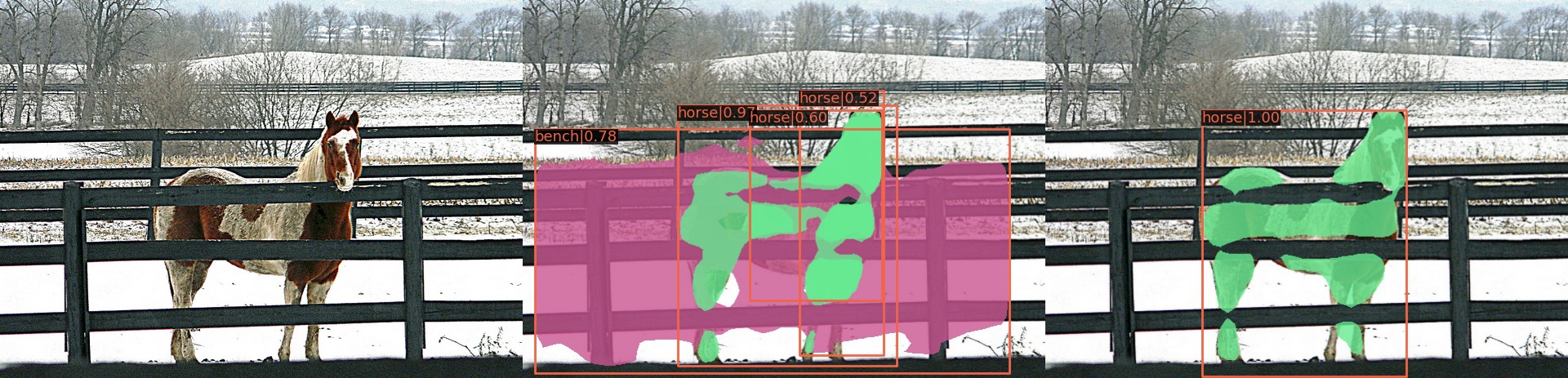}} \\
    \multicolumn{3}{c}{\includegraphics[width=0.45\textwidth]{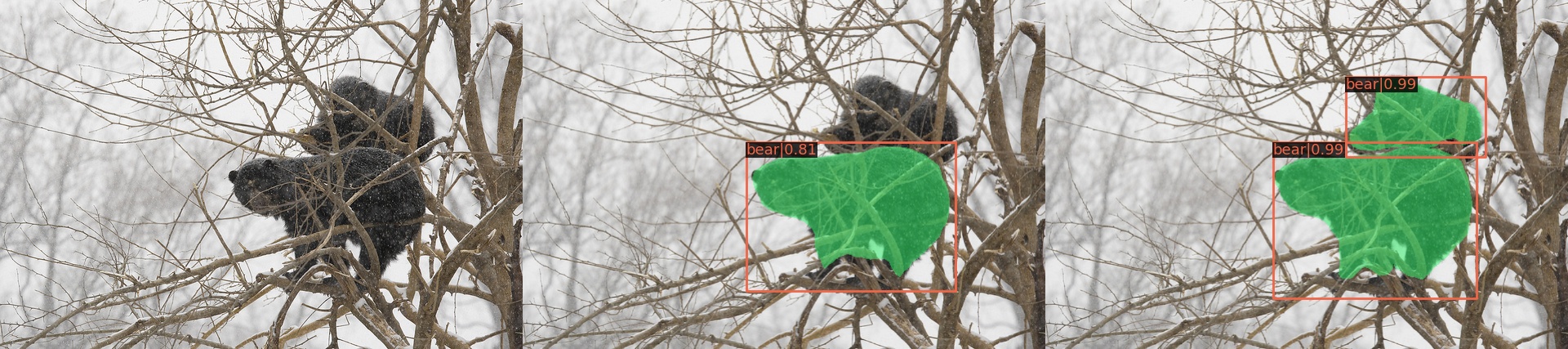}} \\
    \multicolumn{3}{c}{\includegraphics[width=0.45\textwidth]{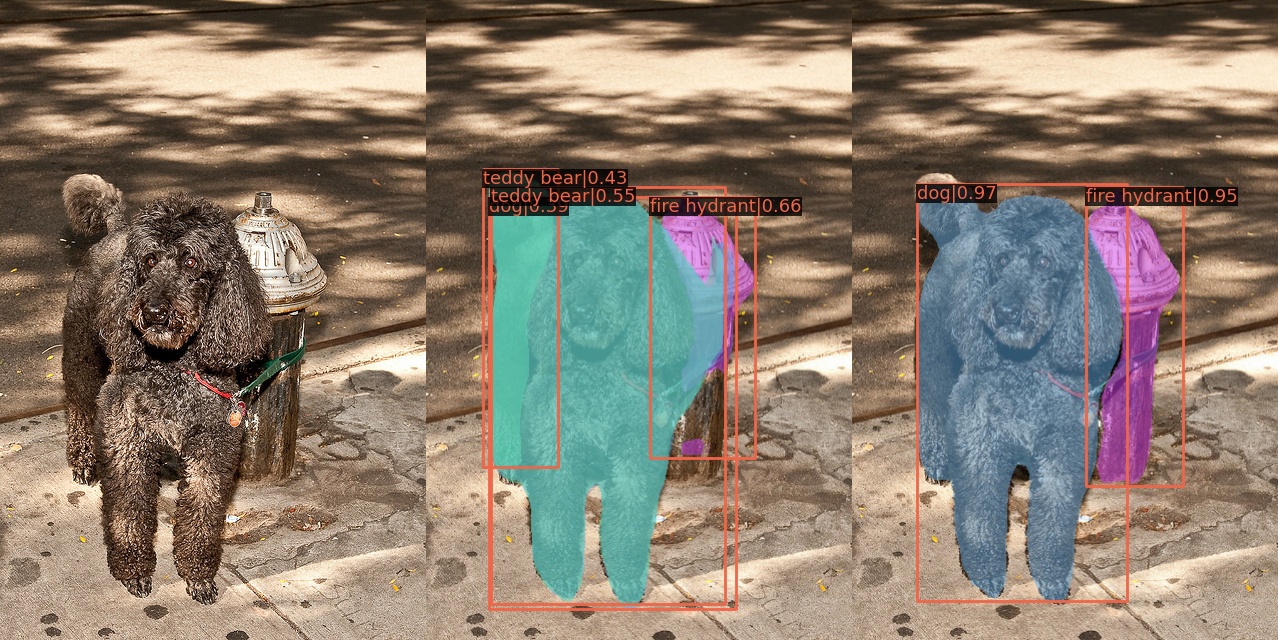}} \\
    Image & Convolution & \textbf{LESA} \\
    \multicolumn{3}{c}{\includegraphics[width=0.45\textwidth]{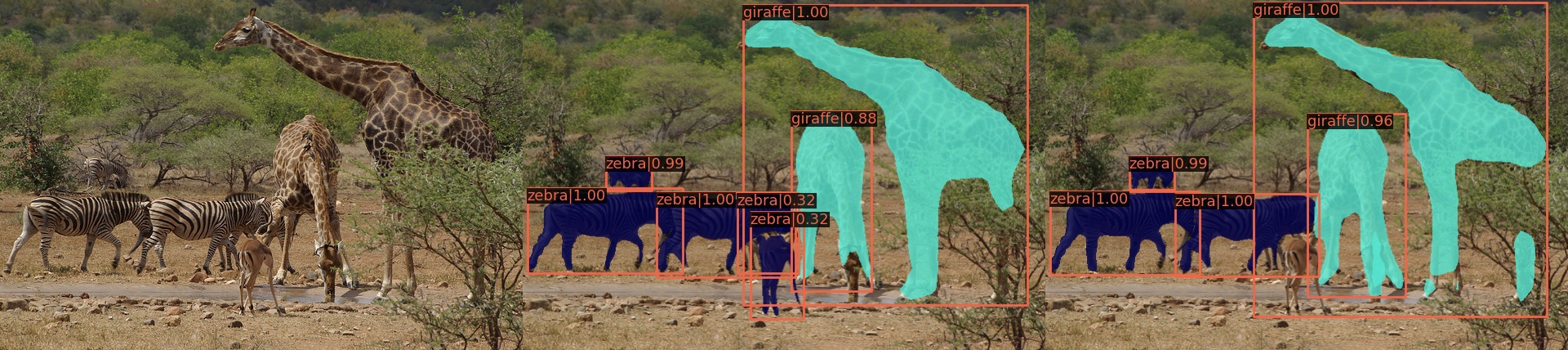}} \\
    \multicolumn{3}{c}{\includegraphics[width=0.45\textwidth]{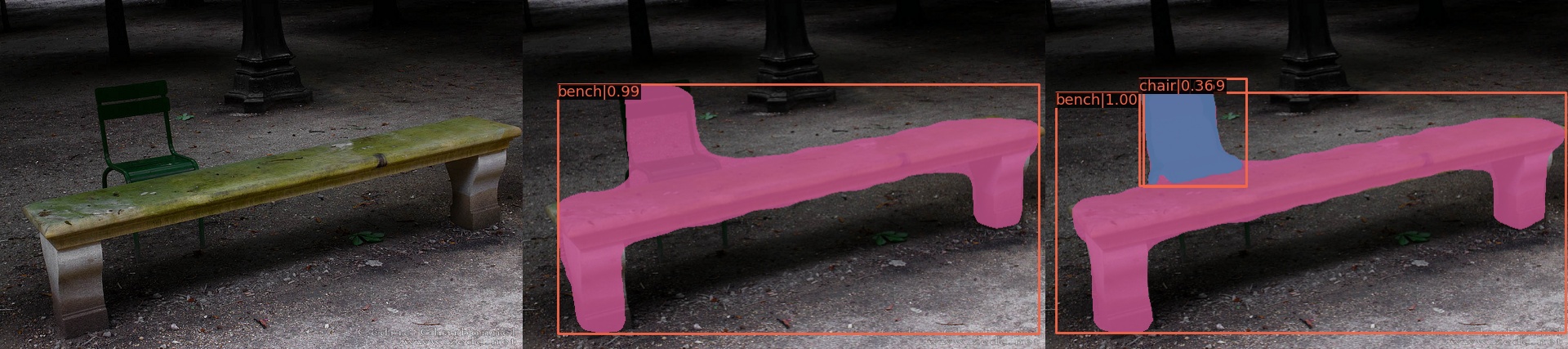}} \\
    \multicolumn{3}{c}{\includegraphics[width=0.45\textwidth]{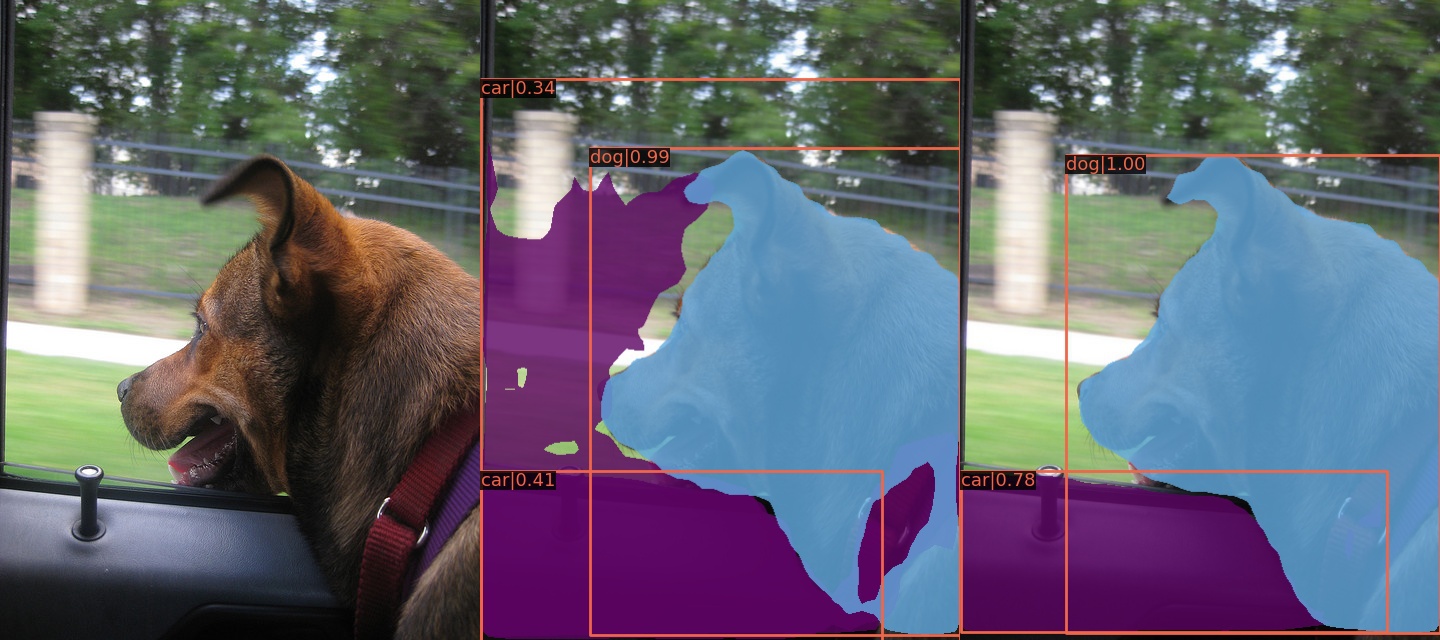}} \\
    \multicolumn{3}{c}{\includegraphics[width=0.45\textwidth]{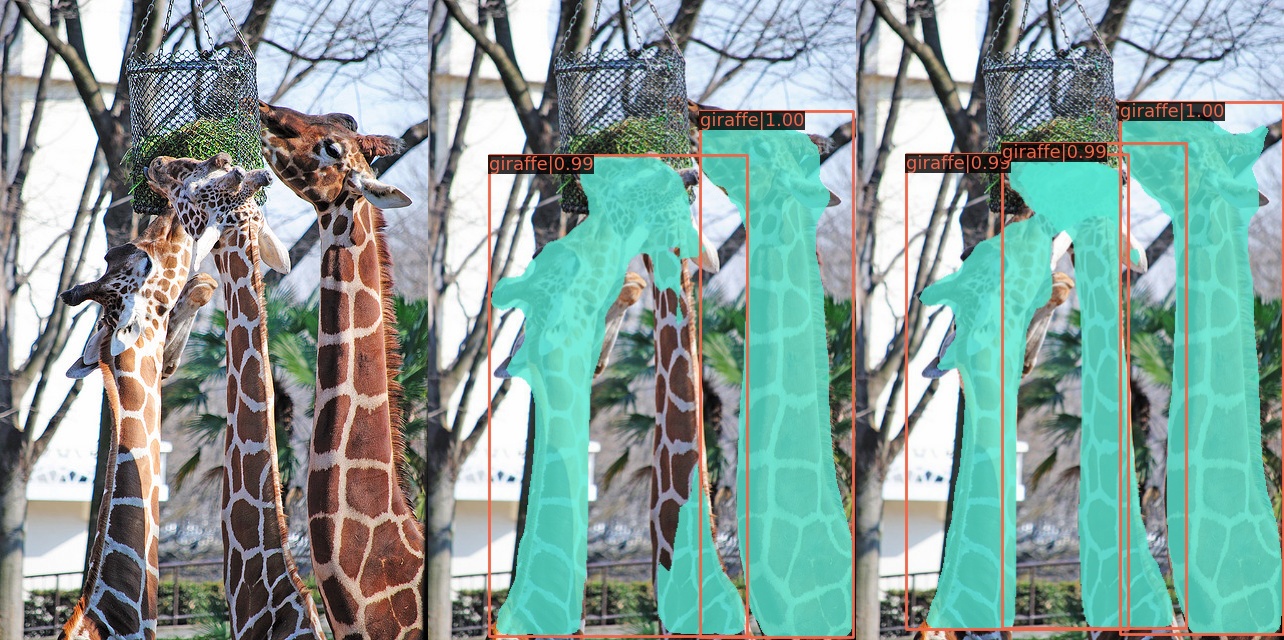}} \\
    Image & Self-Attention & \textbf{LESA} \\
 \end{tabular}
 \captionof{figure}{COCO object detection and instance segmentation.}
 \label{fig: visual comparisons_8}
\end{table}

%% file: 05_ablation_studies.tex
\section{Relationships with other methods}

\noindent$\bullet$\quad{\bf Relationship with other enhanced self-attention methods}
LESA is proposed as a combination mechanism for convolution and self-attention, which can be used with other enhanced self-attention together.

\noindent$\bullet$\quad{\bf Relationships with other merging methods for self-attention and convolution}
\begin{table}[t]
  \centering
  \small
  \begin{tabular}{c|c}
    \toprule
    Categories & Methods \\
    \toprule
    Conv. as position embedding of SA & CVPT; CoaT  \\
    Serial Conn. with Conv. before SA & CvT; \\
    Serial Conn. with Conv. after SA &  LocalViT; PVTv2 \\
    \hline
    \multirow{2}{*}{\textbf{Parallel Conn.}} & AA;  \\ 
    {} & static LESA; LESA \\
    \bottomrule
  \end{tabular}
  \caption{Categories of merging methods for convolution and self-attention(discussed in the related work section). Conn. stands for connection. Approaches in different categories can be used together. As a parallel connection method, LESA is compared with AA (Attention Augmentation) in~\tabref{tab: ablation_imagenet}, and shows the superior performances.}
  \label{tab: category of merging methods}
  \vspace{-0.8em}
\end{table}
As summarized in~\tabref{tab: category of merging methods} and discussed in related work section, there are four categories, and approaches in different ones can be use together. LESA falls into the category of parallel connection. We compare LESA with Attention Augmentation~\cite{bello2019attention} and meanwhile, perform the ablation studies. Static LESA stands for adding the unary and binary terms of LESA without the dynamic fusion module.

The experiments are performed on ImageNet. The setting is the same as the main experiments.
The results are summarized in~\tabref{tab: ablation_imagenet}. It is observed the performance difference between Attention Augmentation and self-attention is marginal. Both Static LESA and LESA show better performances than Attention Augmentation with less parameters. LESA has stronger performance than the static one, demonstrating effectiveness of the dynamic fusion module.

\begin{table}[t]
  \centering
  \small
  \begin{tabular}{c|c|cc}
    \toprule
    \multirow{2}{*}{Operations} & \multirow{2}{*}{Params (M)} & \multicolumn{2}{c}{Accuracy (\%)} \\
    {} & {} & Top-1 & Top-5 \\
    \midrule
    Convolution     & $25.56$ & $76.13$ & $92.86$  \\
    Self-Attention  & $18.06$ & $78.69$ & $94.22$  \\
    AA~\cite{bello2019attention}  & $25.15$ & $78.74$ & $94.23$  \\
    Static LESA  & $20.43$ & $79.12$ & $94.62$  \\
    LESA  & $23.78$ & $79.55$ & $94.79$  \\
    \bottomrule
  \end{tabular}
  \caption{
  Parallel connection methods and ablation studies. The experiments are performed with ResNet50~\cite{he2016deep} on ImageNet. The spatial convolution layers in last two stages are replaced. AA stands for Attention Augmentation~\cite{bello2019attention}, which is another parallel connection method of combining convolution and attention. Static LESA stands for adding the unary and binary terms without the dynamic fusion module. Both Static LESA and LESA achieve better performances than AA with less parameters. The effectiveness of the dynamic fusion module can be proved by the superiority of LESA over Static LESA. 
  }
  \label{tab: ablation_imagenet}
  \vspace{-0.5em}
\end{table}

\newcommand{\figurewidthA}{5.2cm}
\newcommand{\figurewidthB}{2.93cm}
\begin{figure}[t]
\centering
\includegraphics[height=3.8cm]{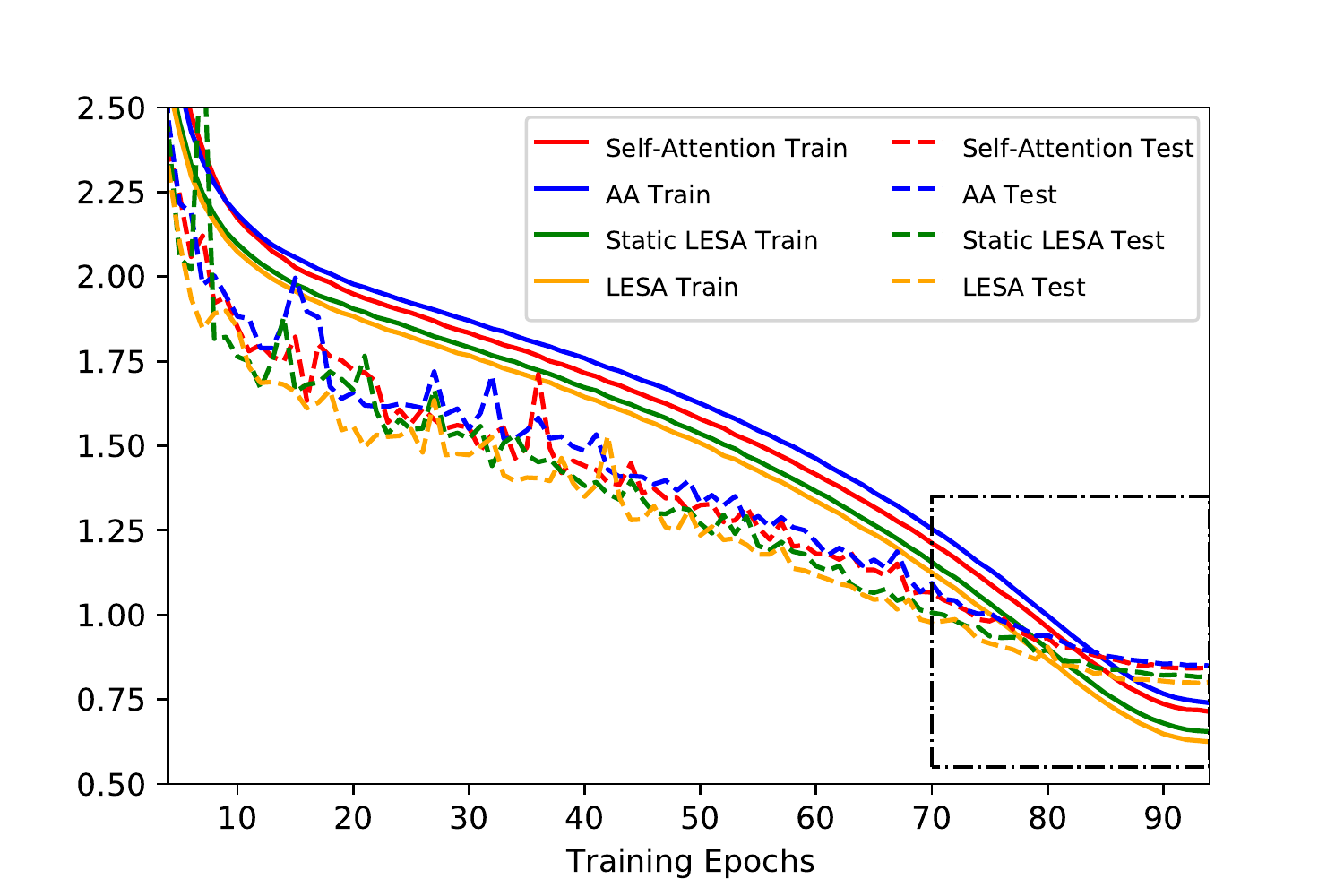}
\includegraphics[height=3.8cm]{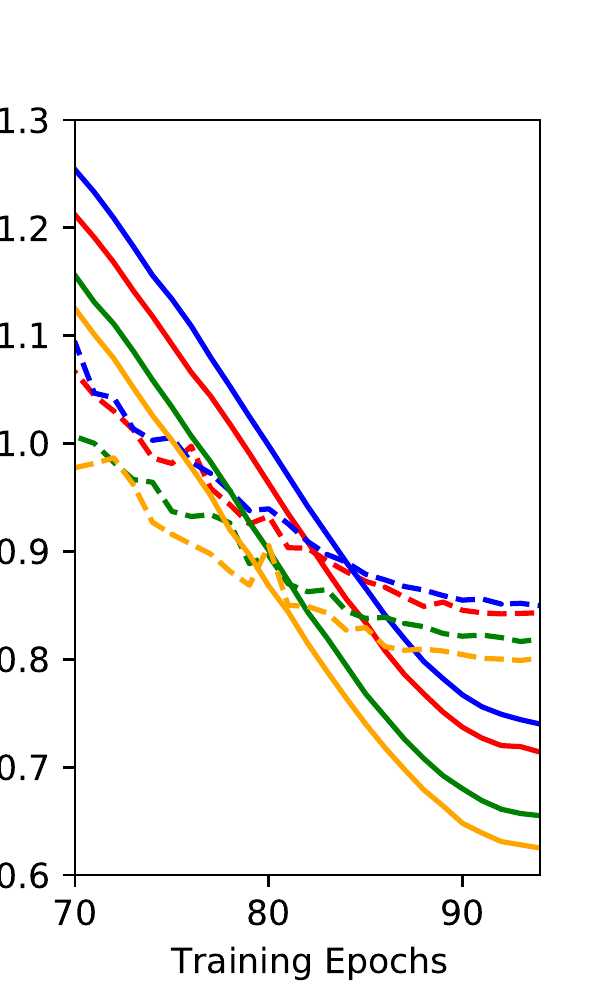}
\caption{Training and testing curves of ablation studies. AA stands for Attention Augmentation~\cite{bello2019attention}. The right figure shows the details in the rectangular region of the left figure.}
\label{Fig:Curves}
\vspace{-0.5em}
\end{figure}

%% file: 06_conclusion.tex
\section{Conclusion}

From the perspective of fully connected Conditional Random Fields (CRFs), we decouple the self-attention into the local and context terms. 
They are the unary and binary terms that are calculated by the queries, keys and values in the attention mechanism. However, there lacks distinction between the local and context cues as they are obtained by using the same set of projection matrices. In addition, we observe the contribution of the local terms is very small which is controlled by the softmax operation. By contrast, the standard Convolutional Neural Networks (CNNs) show excellent performances in various vision tasks and rely solely on the local terms.

In this work, we propose Locally Enhanced Self-Attention (LESA).
First, we enhance the unary term by incorporating it with convolutions.
The multi-head mechanism is realized by using grouped convolution followed by the projection layer. Second, we propose a dynamic fusion module to combine the unary and binary terms with input-dependent relative weights.
We demonstrate the superiority of LESA over the convolution and self-attention baselines in the tasks of ImageNet classification, and COCO object detection and instance segmentation.
\vspace{-1em}

%% file: 07_limitations.tex
\paragraph{Limitations}
LESA \textit{shares} a common limitation with self-attention, which is the large memory consumption. This is due to the large dimensions of the similarity matrix which is computed by the queries and keys and where the softmax operation is applied.
Our future works include designing a LESA that consumes small memory but still has the great power of capturing the context information.
This will also address the common memory issue in other self-attention models.

%% file: main.bbl
\begin{thebibliography}{10}\itemsep=-1pt

\bibitem{bahdanau2014neural}
Dzmitry Bahdanau, Kyunghyun Cho, and Yoshua Bengio.
\newblock Neural machine translation by jointly learning to align and
  translate.
\newblock {\em arXiv preprint arXiv:1409.0473}, 2014.

\bibitem{bello2019attention}
Irwan Bello, Barret Zoph, Ashish Vaswani, Jonathon Shlens, and Quoc~V Le.
\newblock Attention augmented convolutional networks.
\newblock In {\em Proceedings of the IEEE/CVF International Conference on
  Computer Vision}, pages 3286--3295, 2019.

\bibitem{buades2005non}
Antoni Buades, Bartomeu Coll, and J-M Morel.
\newblock A non-local algorithm for image denoising.
\newblock In {\em 2005 IEEE Computer Society Conference on Computer Vision and
  Pattern Recognition (CVPR'05)}, volume~2, pages 60--65. IEEE, 2005.

\bibitem{carion2020end}
Nicolas Carion, Francisco Massa, Gabriel Synnaeve, Nicolas Usunier, Alexander
  Kirillov, and Sergey Zagoruyko.
\newblock End-to-end object detection with transformers.
\newblock In {\em European Conference on Computer Vision}, pages 213--229.
  Springer, 2020.

\bibitem{chen2019hybrid}
Kai Chen, Jiangmiao Pang, Jiaqi Wang, Yu Xiong, Xiaoxiao Li, Shuyang Sun,
  Wansen Feng, Ziwei Liu, Jianping Shi, Wanli Ouyang, et~al.
\newblock Hybrid task cascade for instance segmentation.
\newblock In {\em Proceedings of the IEEE/CVF Conference on Computer Vision and
  Pattern Recognition}, pages 4974--4983, 2019.

\bibitem{chen2019mmdetection}
Kai Chen, Jiaqi Wang, Jiangmiao Pang, Yuhang Cao, Yu Xiong, Xiaoxiao Li,
  Shuyang Sun, Wansen Feng, Ziwei Liu, Jiarui Xu, et~al.
\newblock Mmdetection: Open mmlab detection toolbox and benchmark.
\newblock {\em arXiv preprint arXiv:1906.07155}, 2019.

\bibitem{chen2014semantic}
Liang-Chieh Chen, George Papandreou, Iasonas Kokkinos, Kevin Murphy, and Alan~L
  Yuille.
\newblock Semantic image segmentation with deep convolutional nets and fully
  connected crfs.
\newblock {\em arXiv preprint arXiv:1412.7062}, 2014.

\bibitem{chen2017deeplab}
Liang-Chieh Chen, George Papandreou, Iasonas Kokkinos, Kevin Murphy, and Alan~L
  Yuille.
\newblock Deeplab: Semantic image segmentation with deep convolutional nets,
  atrous convolution, and fully connected crfs.
\newblock {\em IEEE transactions on pattern analysis and machine intelligence},
  40(4):834--848, 2017.

\bibitem{chen2020scaling}
Liang-Chieh Chen, Huiyu Wang, and Siyuan Qiao.
\newblock Scaling wide residual networks for panoptic segmentation.
\newblock {\em arXiv preprint arXiv:2011.11675}, 2020.

\bibitem{chen20182}
Yunpeng Chen, Yannis Kalantidis, Jianshu Li, Shuicheng Yan, and Jiashi Feng.
\newblock $ a^{2} $-nets: Double attention networks.
\newblock {\em arXiv preprint arXiv:1810.11579}, 2018.

\bibitem{chollet2017xception}
Fran{\c{c}}ois Chollet.
\newblock Xception: Deep learning with depthwise separable convolutions.
\newblock In {\em Proceedings of the IEEE conference on computer vision and
  pattern recognition}, pages 1251--1258, 2017.

\bibitem{chu2021conditional}
Xiangxiang Chu, Zhi Tian, Bo Zhang, Xinlong Wang, Xiaolin Wei, Huaxia Xia, and
  Chunhua Shen.
\newblock Conditional positional encodings for vision transformers.
\newblock {\em arXiv preprint arXiv:2102.10882}, 2021.

\bibitem{chung2014empirical}
Junyoung Chung, Caglar Gulcehre, KyungHyun Cho, and Yoshua Bengio.
\newblock Empirical evaluation of gated recurrent neural networks on sequence
  modeling.
\newblock {\em arXiv preprint arXiv:1412.3555}, 2014.

\bibitem{dai2017deformable}
Jifeng Dai, Haozhi Qi, Yuwen Xiong, Yi Li, Guodong Zhang, Han Hu, and Yichen
  Wei.
\newblock Deformable convolutional networks.
\newblock In {\em Proceedings of the IEEE international conference on computer
  vision}, pages 764--773, 2017.

\bibitem{deng2009imagenet}
J. Deng, W. Dong, R. Socher, L.J. Li, K. Li, and L. Fei-Fei.
\newblock Imagenet: A large-scale hierarchical image database.
\newblock In {\em Computer Vision and Pattern Recognition}, 2009.

\bibitem{dong2021attention}
Yihe Dong, Jean-Baptiste Cordonnier, and Andreas Loukas.
\newblock Attention is not all you need: Pure attention loses rank doubly
  exponentially with depth.
\newblock {\em arXiv preprint arXiv:2103.03404}, 2021.

\bibitem{dosovitskiy2020image}
Alexey Dosovitskiy, Lucas Beyer, Alexander Kolesnikov, Dirk Weissenborn,
  Xiaohua Zhai, Thomas Unterthiner, Mostafa Dehghani, Matthias Minderer, Georg
  Heigold, Sylvain Gelly, et~al.
\newblock An image is worth 16x16 words: Transformers for image recognition at
  scale.
\newblock {\em arXiv preprint arXiv:2010.11929}, 2020.

\bibitem{he2017mask}
Kaiming He, Georgia Gkioxari, Piotr Doll{\'a}r, and Ross Girshick.
\newblock Mask r-cnn.
\newblock In {\em Proceedings of the IEEE international conference on computer
  vision}, pages 2961--2969, 2017.

\bibitem{he2016deep}
Kaiming He, Xiangyu Zhang, Shaoqing Ren, and Jian Sun.
\newblock Deep residual learning for image recognition.
\newblock In {\em Proceedings of the IEEE conference on computer vision and
  pattern recognition}, pages 770--778, 2016.

\bibitem{he2016identity}
Kaiming He, Xiangyu Zhang, Shaoqing Ren, and Jian Sun.
\newblock Identity mappings in deep residual networks.
\newblock In {\em European conference on computer vision}, pages 630--645.
  Springer, 2016.

\bibitem{hochreiter1997long}
Sepp Hochreiter and J{\"u}rgen Schmidhuber.
\newblock Long short-term memory.
\newblock {\em Neural computation}, 9(8):1735--1780, 1997.

\bibitem{howard2017mobilenets}
Andrew~G Howard, Menglong Zhu, Bo Chen, Dmitry Kalenichenko, Weijun Wang,
  Tobias Weyand, Marco Andreetto, and Hartwig Adam.
\newblock Mobilenets: Efficient convolutional neural networks for mobile vision
  applications.
\newblock {\em arXiv preprint arXiv:1704.04861}, 2017.

\bibitem{hu2019local}
Han Hu, Zheng Zhang, Zhenda Xie, and Stephen Lin.
\newblock Local relation networks for image recognition.
\newblock In {\em Proceedings of the IEEE/CVF International Conference on
  Computer Vision}, pages 3464--3473, 2019.

\bibitem{hu2018squeeze}
Jie Hu, Li Shen, and Gang Sun.
\newblock Squeeze-and-excitation networks.
\newblock In {\em Proceedings of the IEEE conference on computer vision and
  pattern recognition}, pages 7132--7141, 2018.

\bibitem{huang2017densely}
Gao Huang, Zhuang Liu, Laurens Van Der~Maaten, and Kilian~Q Weinberger.
\newblock Densely connected convolutional networks.
\newblock In {\em Proceedings of the IEEE conference on computer vision and
  pattern recognition}, pages 4700--4708, 2017.

\bibitem{ioffe2015batch}
Sergey Ioffe and Christian Szegedy.
\newblock Batch normalization: Accelerating deep network training by reducing
  internal covariate shift.
\newblock In {\em International conference on machine learning}, pages
  448--456. PMLR, 2015.

\bibitem{krizhevsky2012imagenet}
Alex Krizhevsky, Ilya Sutskever, and Geoffrey~E Hinton.
\newblock Imagenet classification with deep convolutional neural networks.
\newblock {\em Advances in neural information processing systems},
  25:1097--1105, 2012.

\bibitem{li2021localvit}
Yawei Li, Kai Zhang, Jiezhang Cao, Radu Timofte, and Luc Van~Gool.
\newblock Localvit: Bringing locality to vision transformers.
\newblock {\em arXiv preprint arXiv:2104.05707}, 2021.

\bibitem{lin2017feature}
Tsung-Yi Lin, Piotr Doll{\'a}r, Ross Girshick, Kaiming He, Bharath Hariharan,
  and Serge Belongie.
\newblock Feature pyramid networks for object detection.
\newblock In {\em Proceedings of the IEEE conference on computer vision and
  pattern recognition}, pages 2117--2125, 2017.

\bibitem{lin2014microsoft}
Tsung-Yi Lin, Michael Maire, Serge Belongie, James Hays, Pietro Perona, Deva
  Ramanan, Piotr Doll{\'a}r, and C~Lawrence Zitnick.
\newblock Microsoft coco: Common objects in context.
\newblock In {\em European conference on computer vision}, pages 740--755.
  Springer, 2014.

\bibitem{liu2021swin}
Ze Liu, Yutong Lin, Yue Cao, Han Hu, Yixuan Wei, Zheng Zhang, Stephen Lin, and
  Baining Guo.
\newblock Swin transformer: Hierarchical vision transformer using shifted
  windows.
\newblock {\em arXiv preprint arXiv:2103.14030}, 2021.

\bibitem{nesterov1983method}
Yurii~E Nesterov.
\newblock A method for solving the convex programming problem with convergence
  rate o (1/k\^{} 2).
\newblock In {\em Dokl. akad. nauk Sssr}, volume 269, pages 543--547, 1983.

\bibitem{papandreou2015modeling}
George Papandreou, Iasonas Kokkinos, and Pierre-Andr{\'e} Savalle.
\newblock Modeling local and global deformations in deep learning: Epitomic
  convolution, multiple instance learning, and sliding window detection.
\newblock In {\em Proceedings of the IEEE conference on computer vision and
  pattern recognition}, pages 390--399, 2015.

\bibitem{paszke2019pytorch}
Adam Paszke, Sam Gross, Francisco Massa, Adam Lerer, James Bradbury, Gregory
  Chanan, Trevor Killeen, Zeming Lin, Natalia Gimelshein, Luca Antiga, et~al.
\newblock Pytorch: An imperative style, high-performance deep learning library.
\newblock {\em arXiv preprint arXiv:1912.01703}, 2019.

\bibitem{perronnin2010improving}
Florent Perronnin, Jorge S{\'a}nchez, and Thomas Mensink.
\newblock Improving the fisher kernel for large-scale image classification.
\newblock In {\em European conference on computer vision}, pages 143--156.
  Springer, 2010.

\bibitem{prince2012computer}
Simon~JD Prince.
\newblock {\em Computer vision: models, learning, and inference}.
\newblock Cambridge University Press, 2012.

\bibitem{qiao2020detectors}
Siyuan Qiao, Liang-Chieh Chen, and Alan Yuille.
\newblock Detectors: Detecting objects with recursive feature pyramid and
  switchable atrous convolution.
\newblock {\em arXiv preprint arXiv:2006.02334}, 2020.

\bibitem{ramachandran2019stand}
Prajit Ramachandran, Niki Parmar, Ashish Vaswani, Irwan Bello, Anselm Levskaya,
  and Jonathon Shlens.
\newblock Stand-alone self-attention in vision models.
\newblock {\em arXiv preprint arXiv:1906.05909}, 2019.

\bibitem{russakovsky2015imagenet}
Olga Russakovsky, Jia Deng, Hao Su, Jonathan Krause, Sanjeev Satheesh, Sean Ma,
  Zhiheng Huang, Andrej Karpathy, Aditya Khosla, Michael Bernstein, et~al.
\newblock Imagenet large scale visual recognition challenge.
\newblock {\em International journal of computer vision}, 115(3):211--252,
  2015.

\bibitem{sanchez2011high}
Jorge S{\'a}nchez and Florent Perronnin.
\newblock High-dimensional signature compression for large-scale image
  classification.
\newblock In {\em CVPR 2011}, pages 1665--1672. IEEE, 2011.

\bibitem{shaw2018self}
Peter Shaw, Jakob Uszkoreit, and Ashish Vaswani.
\newblock Self-attention with relative position representations.
\newblock {\em arXiv preprint arXiv:1803.02155}, 2018.

\bibitem{simonyan2014very}
Karen Simonyan and Andrew Zisserman.
\newblock Very deep convolutional networks for large-scale image recognition.
\newblock {\em arXiv preprint arXiv:1409.1556}, 2014.

\bibitem{srinivas2021bottleneck}
Aravind Srinivas, Tsung-Yi Lin, Niki Parmar, Jonathon Shlens, Pieter Abbeel,
  and Ashish Vaswani.
\newblock Bottleneck transformers for visual recognition.
\newblock {\em arXiv preprint arXiv:2101.11605}, 2021.

\bibitem{sutskever2013importance}
Ilya Sutskever, James Martens, George Dahl, and Geoffrey Hinton.
\newblock On the importance of initialization and momentum in deep learning.
\newblock In {\em International conference on machine learning}, pages
  1139--1147. PMLR, 2013.

\bibitem{szegedy2015going}
Christian Szegedy, Wei Liu, Yangqing Jia, Pierre Sermanet, Scott Reed, Dragomir
  Anguelov, Dumitru Erhan, Vincent Vanhoucke, and Andrew Rabinovich.
\newblock Going deeper with convolutions.
\newblock In {\em Proceedings of the IEEE conference on computer vision and
  pattern recognition}, pages 1--9, 2015.

\bibitem{tan2019efficientnet}
Mingxing Tan and Quoc Le.
\newblock Efficientnet: Rethinking model scaling for convolutional neural
  networks.
\newblock In {\em International Conference on Machine Learning}, pages
  6105--6114. PMLR, 2019.

\bibitem{touvron2020training}
Hugo Touvron, Matthieu Cord, Matthijs Douze, Francisco Massa, Alexandre
  Sablayrolles, and Herv{\'e} J{\'e}gou.
\newblock Training data-efficient image transformers \& distillation through
  attention.
\newblock {\em arXiv preprint arXiv:2012.12877}, 2020.

\bibitem{vaswani2017attention}
Ashish Vaswani, Noam Shazeer, Niki Parmar, Jakob Uszkoreit, Llion Jones,
  Aidan~N Gomez, Lukasz Kaiser, and Illia Polosukhin.
\newblock Attention is all you need.
\newblock {\em arXiv preprint arXiv:1706.03762}, 2017.

\bibitem{wan2021high}
Ziyu Wan, Jingbo Zhang, Dongdong Chen, and Jing Liao.
\newblock High-fidelity pluralistic image completion with transformers.
\newblock {\em arXiv preprint arXiv:2103.14031}, 2021.

\bibitem{wang2020max}
Huiyu Wang, Yukun Zhu, Hartwig Adam, Alan Yuille, and Liang-Chieh Chen.
\newblock Max-deeplab: End-to-end panoptic segmentation with mask transformers.
\newblock {\em arXiv preprint arXiv:2012.00759}, 2020.

\bibitem{wang2020axial}
Huiyu Wang, Yukun Zhu, Bradley Green, Hartwig Adam, Alan Yuille, and
  Liang-Chieh Chen.
\newblock Axial-deeplab: Stand-alone axial-attention for panoptic segmentation.
\newblock In {\em European Conference on Computer Vision}, pages 108--126.
  Springer, 2020.

\bibitem{wang2021pvtv2}
Wenhai Wang, Enze Xie, Xiang Li, Deng-Ping Fan, Kaitao Song, Ding Liang, Tong
  Lu, Ping Luo, and Ling Shao.
\newblock Pvtv2: Improved baselines with pyramid vision transformer.
\newblock {\em arXiv preprint arXiv:2106.13797}, 2021.

\bibitem{wang2018non}
Xiaolong Wang, Ross Girshick, Abhinav Gupta, and Kaiming He.
\newblock Non-local neural networks.
\newblock In {\em Proceedings of the IEEE conference on computer vision and
  pattern recognition}, pages 7794--7803, 2018.

\bibitem{wu2021cvt}
Haiping Wu, Bin Xiao, Noel Codella, Mengchen Liu, Xiyang Dai, Lu Yuan, and Lei
  Zhang.
\newblock Cvt: Introducing convolutions to vision transformers.
\newblock {\em arXiv preprint arXiv:2103.15808}, 2021.

\bibitem{wu2020lite}
Zhanghao Wu, Zhijian Liu, Ji Lin, Yujun Lin, and Song Han.
\newblock Lite transformer with long-short range attention.
\newblock {\em arXiv preprint arXiv:2004.11886}, 2020.

\bibitem{xie2017aggregated}
Saining Xie, Ross Girshick, Piotr Doll{\'a}r, Zhuowen Tu, and Kaiming He.
\newblock Aggregated residual transformations for deep neural networks.
\newblock In {\em Proceedings of the IEEE conference on computer vision and
  pattern recognition}, pages 1492--1500, 2017.

\bibitem{xu2021co}
Weijian Xu, Yifan Xu, Tyler Chang, and Zhuowen Tu.
\newblock Co-scale conv-attentional image transformers.
\newblock {\em arXiv preprint arXiv:2104.06399}, 2021.

\bibitem{zagoruyko2016wide}
Sergey Zagoruyko and Nikos Komodakis.
\newblock Wide residual networks.
\newblock {\em arXiv preprint arXiv:1605.07146}, 2016.

\bibitem{zhao2020exploring}
Hengshuang Zhao, Jiaya Jia, and Vladlen Koltun.
\newblock Exploring self-attention for image recognition.
\newblock In {\em Proceedings of the IEEE/CVF Conference on Computer Vision and
  Pattern Recognition}, pages 10076--10085, 2020.

\bibitem{zhu2019deformable}
Xizhou Zhu, Han Hu, Stephen Lin, and Jifeng Dai.
\newblock Deformable convnets v2: More deformable, better results.
\newblock In {\em Proceedings of the IEEE/CVF Conference on Computer Vision and
  Pattern Recognition}, pages 9308--9316, 2019.

\end{thebibliography}
